\documentclass[11pt]{article}

\usepackage[T1]{fontenc}
\usepackage[english]{babel}
\usepackage{amsmath,amsthm,amssymb,amsfonts}
\usepackage{xcolor}
\usepackage{graphicx}
\usepackage{algorithm}
\usepackage{algpseudocode}
\usepackage{tikz}
\usepackage[margin=1in]{geometry}
\usepackage{setspace}
\usepackage[nopatch=footnote]{microtype}
\usepackage{indentfirst}
\usepackage{soul}
\usepackage[numbers]{natbib}
\usepackage[colorlinks=true,linkcolor=blue,urlcolor=blue,citecolor=blue]{hyperref}
\usepackage[nameinlink]{cleveref}
\usepackage{booktabs}
\usepackage{rotating}
\usepackage{multirow}
\usepackage{float}

\renewcommand{\epsilon}{\varepsilon}

\title{AutoOnboarder: Autonomously Onboarding LLM Agents}

\author{
Tanay Arora\\
Research Science Institute, Massachusetts Institute of Technology\\
CliGenAILab, Boston Children's Hospital\\
\and
Cong Liu\\
CliGenAILab, Boston Children's Hospital
}

\date{August 2026}

\begin{document}

\maketitle

\begin{abstract}
 Agentic large language models have been documented to lose reasoning capability when placed inside a harness, the software that mediates their observations and actions. To reconcile these losses, we introduce AutoOnboarder, an autonomous system that adapts an agent to a task the way a trainer onboards a new employee. It places the agent under the supervision of an intelligent ``Onboarder'' coding agent, which iteratively tailors its foundation model, prompt, and harness source code to simultaneously improve performance and decrease cost. Like a real onboarding process, the Onboarder is only given tens of examples, and all evaluation data is isolated on a separate virtual machine to prevent leakage. We report the lessons learned from building and testing AutoOnboarder on genomic-sequencing prior authorization form submission. Autonomously onboarded agents are able to match or exceed the performance of a Gemini 3.5 Flash baseline at up to 29$\times$ lower costs. However, these agents remain imperfect, and the Onboarding process varies greatly in terms of final performance. We conclude that future Onboarding systems require a robust environment for the Onboarder, including a structured thought interface, (controlled) internet access, and a wide model search space.  
\end{abstract}

\section{Introduction}

\subsection{The Harness Bottleneck}

In today's world, large language models (LLMs) are increasingly used as agents. Rather than simply answering questions, they autonomously complete tasks by planning and interacting with tools. This interaction is supported by a harness: a program that supplies a model with its context, mediates its access to tools, interprets its outputs as actions, and decides when a task is finished.

Unfortunately, these harnesses are also liabilities. An agent spends a significant part of its limited context interacting with its harness, and that overhead competes with the model's reasoning~\cite{levy_same_2024, shi_distracted_2023, liu_lost_2024}.

To illustrate this flaw, Nie et al. design a sandbox -- a simulation of a real task environment -- for whole-exome and whole-genome sequencing prior-authorization form submission~\cite{nie_hard_2026}. In their work, when frontier models were asked directly whether an artificial patient record contained an error, they flagged errors correctly 91\% of the time. The same models, when placed under a browser-use harness inside of their sandbox, nearly universally failed to recognize errors (\(<\)~0.5\%) and instead submitted deficient forms \cite{nie_hard_2026}. The act of interacting with a harness correlated with worse critical thinking capability for these agents. This work focuses on developing a method to automatically adapt an agent and its harness to a task, with the aim of recovering lost reasoning capability.

\subsection{Related Work}

Here, adapting an agent means optimizing it to perform best on a target task. Current methods for adapting agents can be divided into four main families (\Cref{tab:related}). 

\begin{table}[tb]
\centering
\begin{singlespace}
\small
\begin{tabular}{lcccccl}
\toprule
& \multicolumn{4}{c}{May change} & & \\
\cmidrule(lr){2-5}
Family & Weights & Prompt & Model & Harness & \shortstack{Cost\\objective} \\
\midrule
Finetuning      & \checkmark & --         & --         & --         & -- \\
Prompt-adaptation & --         & \checkmark & --         & --          & -- \\
Interface design   & --         & \checkmark & --         & \checkmark            & -- \\
Agents improving agents   & --         & \checkmark & -- & \checkmark &  -- \\
\textbf{Onboarding } & --      & \checkmark & \checkmark & \checkmark           & \checkmark \\
\bottomrule
\end{tabular}
\end{singlespace}
\caption{Scopes of agent-adaptation families.}
\label{tab:related}
\end{table}

The first family updates model weights directly: finetuning and Low-Rank Adaptation (LoRA)~\cite{hu_lora_2021} tune model parameters on labeled task data and reinforcement learning approaches tune them based on policy gradients or value-based approaches. The methods in this family require thousands of examples or expensive trajectory sampling, and neither change what the deployed agent has access to during inference.

The second family adapts the prompt while the model and harness remain fixed. Search-based optimizers generate and select candidate instructions (APE~\cite{zhou_ape_2022}), feed previously scored solutions back to an optimizer LLM (OPRO~\cite{yang_large_2023}), or edit prompts along natural-language ``gradients'' (ProTeGi~\cite{pryzant_automatic_2023}, DSPy~\cite{khattab_dspy_2024}, TextGrad~\cite{yuksekgonul_optimizing_2025}). An experiential branch distills a few of the agent's own episodes into lessons: self-reflections (Reflexion~\cite{shinn_reflexion_2023}, ExpeL~\cite{zhao_expel_2024}, Learn-by-Interact~\cite{su_learnbyinteract_2025}), environment rule manuals (AutoManual~\cite{chen_automanual_2024}), and state-conditioned guidelines (AutoGuide~\cite{fu_autoguide_2024}). These approaches adapt agents with only a few examples, but everything outside the prompt and memory remains fixed. 

The third family redesigns the harness itself: SWE-agent's purpose-built agent-computer interface substantially improves software-engineering agents~\cite{yang_swe_2024}, and AgentOccam gains similarly on web tasks by aligning observation and action spaces with the underlying LLM~\cite{yang_occam_2024}. However, these works redesign interfaces by hand rather than autonomously, which can make adaptation tedious.

The most similar work to ours lets agents improve themselves or other agents. In ADAS, a meta-agent programs new agents in code, including edits to prompts and tools, and archives the best designs~\cite{shengran_adas_2024}; the Darwin Gödel Machine iteratively rewrites its own codebase, keeping changes that improve coding benchmarks~\cite{zhang_darwin_2025}. Both optimize a large benchmark score over hundreds of evaluations and iterations, and neither carries an explicit cost objective for the agents it produces.

None of these families matches how a capable human is onboarded onto a task. When a new employee enters a job, they are often assigned to an onboarding course where a trainer walks them through a few representative cases, the varying quirks of the tools, and frameworks on how to think about the task. This process of onboarding uses only tens or hundreds of examples, and adjusts everything the worker can see (as opposed to only the foundation model weights or only the prompt). To our knowledge, no method today fully plays this role for an agent, especially while simultaneously reducing costs.

\subsection{AutoOnboarder}

In this work, we develop \textit{AutoOnboarder}: an autonomous, agentic system for ``onboarding'' other agents. The system pairs two agents: an \textit{Executor}, the agent being adapted to a task, and an \textit{Onboarder}, an intelligent coding agent that adapts it. Given a small Onboarding set (only tens of cases), the Onboarder optimizes the Executor and its harness for both cost and performance. To best emulate an onboarding environment, where an employee is meticulously walked through a few representative cases, we give the Onboarder access to virtually everything the Executor sees, while hiding all evaluation (``real-world'') data on a separate virtual machine to avoid data leakage. We hypothesized that Onboarding could recover the critical thinking capability documented by Nie et al.~\cite{nie_hard_2026}, while simultaneously improving cost-effectiveness. Our results are mixed, and we report them as lessons learned rather than as a finished method. In this paper, we contribute (i) the AutoOnboarder system and its two-machine, leakage-controlled evaluation protocol; (ii) a reliability study of the evaluation itself, covering judge agreement and breaking down sources of variance; (iii) an evaluation of the effectiveness of including the harness in Onboarding; (iv) an observational study of generalization to unseen error classes; and (v) practical lessons for building Onboarding systems, drawn from our experimental logs. Because task evaluation is expensive and time-consuming, each experiment is run once and repeated at most three times; throughout this paper, we report results as directional evidence and avoid statistical tests.

\section{Methodology}

\subsection{The Onboarding Loop}
\label{sec:loop}

The Onboarder is given a set number of iterations and a per-case cost goal, \(\beta\), and must produce the best Executor candidate that it can near the goal. For each iteration, the Onboarder is instructed to implement a ``single, identifiable idea'' in exactly one of its three levers (defined below). We intentionally keep this instruction vague and let the Onboarder interpret it however it likes, so its changes are precisely what it deems necessary. We never provide it with examples of changes it can make, so it must devise all changes by its own rationale. The three levers that the Onboarder has access to include:

\begin{algorithm}[bt]
\caption{The Onboarding loop}\label{alg:loop}
\begin{algorithmic}[1]
\Require Onboarding set $\mathcal{D}$, cost goal $\beta$, iteration count $T$, initial config. $c_0$
\For{$t = 1, \dots, T$}
  \State implement one ``idea'' to one lever of $c_{t-1}$, giving $c'$
  \State run the Executor with $c'$ on all of $\mathcal{D}$; record logs and per-case cost
  \State send episodes to the Trainer; receive scores and per-case feedback
  \State $c_t \gets c'$ or $c_{t-1}$, at the Onboarder's judgement

\EndFor
\State \Return one kept configuration, nominated for validation
\end{algorithmic}
\end{algorithm}
\begin{itemize}
  \item Prompt: the Executor's task-prompt template, which is wrapped by the harness before being fed into the model. An ``idea'' here can be any edit to this template.
  \item Foundation Model: any OpenRouter model identifier. OpenRouter is a unified router to hundreds of different LLMs from various providers by name, so an ``idea'' here likely swaps one model identifier to another (e.g., `google/gemini-3.5-flash' to `openai/gpt-5.6-luna'). The Onboarder is provided with a helper script to inspect affordable options.
  \item Harness: the Executor's harness source code. For our experiments, the harness refers to the browser-use harness \cite{browser_use2024}, which is installed in the Onboarder's local directory. The Onboarder can edit anything in this harness (e.g., history logic, system prompts, action definitions). An ``idea'' here might range from cutting the action registry to programmatically scraping patient records directly into the Executor's persistent memory.
\end{itemize}

In each iteration, after making an isolated change, the Onboarder reevaluates the Executor on the Onboarding data; examines its logs, score, and feedback; and chooses to keep or revert its change based on its own judgement (\Cref{alg:loop}). At the end of its iterations, the Onboarder is told to nominate exactly one iteration, whichever it determines to be its best, for evaluation. The Onboarder is also given a hard per-case cost ceiling, computed as \(1.5\beta\). If any iteration is above this ceiling when evaluated on the Onboarding data, it cannot be nominated for validation. Additionally, to test the Onboarder's ability to reduce costs autonomously, the initial configuration of the Executor sits far above its cost ceiling.

\subsection{Evaluation Task}
\label{sec:task}

\begin{figure}[tb]
    \centering
    \includegraphics[width=1\linewidth]{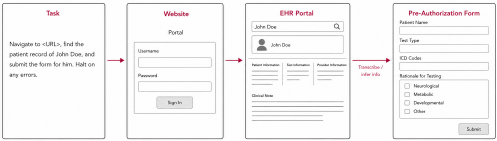}
    \caption{Heavily simplified visualization of the Executor's evaluation task. }
    \label{fig:executor_workflow}
\end{figure}

We instantiate AutoOnboarder on whole-exome and whole-genome sequencing prior authorization, which our laboratory previously used as a sandbox~\cite{nie_hard_2026}. Prior-authorization is an insurer-initiated check of medical necessity, and for sequencing it is a documented bottleneck to rare-disease diagnosis. Roughly one in five requests are denied~\cite{smith_priorauth_2021} and physicians report losing about thirteen hours a week to the paperwork~\cite{ama_priorauth_2022}. The sandbox pairs a simulated web physician portal with a synthetic electronic health record (EHR) system, and each dataset sample consists of exactly one patient profile in this system. Some patient profiles are deficient, seeded with irreparable errors that should trigger a halt rather than a submission (\Cref{app:b_error_classes}). Given only a patient's name, the Executor must log in, search for and read the patient's EHR record, and either complete the form or halt and report any issues (\Cref{fig:executor_workflow}). Completing the form requires both transcribing information from EHR record fields and inferring entries from the attached clinical note.

Scoring is approached in a binary manner. A \textit{success} corresponds to (i) halting on a deficient case and providing the correct reason or (ii) submitting a valid case with perfect form-field accuracy. Any other situation is considered a \textit{failure}, whether it be halting with an incorrect reason, halting on a valid case, a technical failure, submitting a deficient case, or imperfect form-field accuracy on a valid submission. An Executor that takes 55 steps (LLM calls) without finishing is automatically stopped and the episode is reported as a technical failure. The score, \(S\), of a set of cases is calculated as

\begin{equation}
\label{eq:score} 
s(\text{case}) =
\begin{cases}
0 &\textit{failure}\\
1 & \textit{success}
\end{cases} \ ,
\qquad \quad
S = \tfrac{1}{2}\,\bar{s}_{\text{ valid}} + \tfrac{1}{2}\,\bar{s}_{\text{ deficient}},
\end{equation}

\noindent where \(\bar{s}_{\text{class}}\) is the mean over that class; the score weights the valid and deficient classes equally. A small LLM is tasked with deciding whether a reason provided is correct or incorrect by comparing it to the case's ground truth issue description. We also define \textrm{MFA} as the mean form-field accuracy over all valid cases -- where valid cases that are not submitted contribute 0\% -- and \textrm{CH} as the percentage of deficient cases correctly halted. 

For this study, we make the underlying dataset of the Hard-to-Halt study~\cite{nie_hard_2026} more difficult by expanding it from five fixed error classes to 23, spread over three categories: defects in to-be-transcribed EHR records, defects in clinical notes that make the requested sequencing illogical, and cross-field contradictions between the clinical note and other EHR record fields. The validation set holds 180 cases, and Onboarding sets are sampled from its distribution. Exact class counts are listed in \Cref{app:b_error_classes}; there are 141 (78.3\%) deficient cases and 39 (21.7\%) valid cases. In any given experiment, we use a dataset of 200 total cases in a 10/90 split (20 Onboarding, 180 validation). Because the expanded number of error classes is greater than the number of Onboarding samples, this dataset design forces the Onboarder to avoid overfitting the Executor to the Onboarding data.

\subsection{Information Boundaries}
\label{sec:arch}

\begin{figure}[tb]
  \centering
    \includegraphics[
    width=0.9\linewidth,
    height=0.32\textheight,
    keepaspectratio=false]{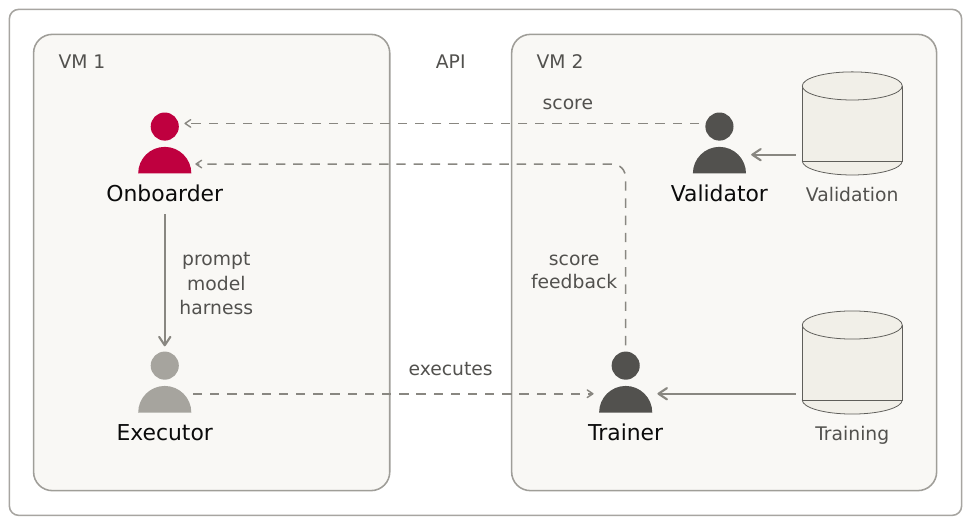}
  \caption{The two-machine architecture. }
  \label{fig:arch}
\end{figure}

\noindent We split the system across two virtual machines to enforce barriers on the Onboarder and prevent information leakage (\Cref{fig:arch}).

\textbf{Virtual Machine 1} hosts the agents: an Onboarder and an Executor. The Onboarder is a Coding Agent (e.g., Claude Code, OpenAI Codex) that has full control of the three levers, direct JSON-level access to sanitized copies of the Onboarding cases (stripped of all error class information), the Executor's step-by-step logs, and the Trainer's feedback. Like a real onboarding trainer, the Onboarder has access to a great deal of information, and it is up to its own reasoning to extract this information without overfitting the Executor. The Onboarder is also given a helper script to inspect the content of the sandbox website.

\textbf{Virtual Machine 2} holds all ground truth data and exposes two API endpoints: a Trainer and a Validator. The Trainer sends and scores Onboarding batches exclusively and returns a softened version of the score and per-case textual feedback. This softened score, \(S_{\mathrm{soft}}\), computes \(s_{\text{valid}} = \textrm{MFA}\), rather than \(s_{\text{valid}} =\) (fraction of successes). This design is used because it provides a more clear heuristic for success to the Onboarder; the Onboarder is still told that validation will be binary. The Trainer never reveals the error classes of the dataset (\Cref{app:b_error_classes}), but it does indicate whether a provided reason is correct in its feedback. On the other hand, the Validator 
sends and scores validation batches exclusively, is called only by the human operator, and returns nothing but the numerical score; per-case feedback stays on VM2.

To prevent the Onboarder from gaming the task and examining hidden validation data, many structural boundaries are in place. It has no internet access, and all communication across machines is in the form of token-authenticated local API calls. The Executor runs each browser-use episode in a locally-staged sandbox that signs all submissions with HMAC, so the Onboarder cannot fabricate submissions. Filesystem permissions restrict the Onboarder's write access to its three levers and logging files, keep instruction files read-only, and deny anything that would reveal hidden paths or generation code (the data creation pipeline, the staged app). A small set of less severe rules (e.g., never trigger validation scoring, never bypass cost restrictions, never view other experiments' git branches) is enforced implicitly by a written ``contract'' presented to the Onboarder, with compliance verified manually post-hoc.

\section{Experiments}
\label{sec:design}

\subsection{Experimental Configuration}
\label{sec:protocol}

In our experiments, we use Claude Code running Opus 4.8 at max effort~\cite{anthropic_claudecode_2025} as the Onboarder, with a budget of 12 iterations. We use OpenAI's gpt-oss-120B as the foundation model for the judge that determines correctness of responses. Per-case cost is calculated via the USD charge that OpenRouter reports for the episode's API calls. Every run starts from a large baseline: Gemini 3.5 Flash with the prompt from Nie et al.~\cite{nie_hard_2026} and an unmodified harness (\$0.94 per case). We run Onboarding at three per-case cost goals, \(\beta \in \{ \$0.1, \$0.04, \$0.01\}\). These per-case cost goals are chosen primarily due to cost constraints, though pilot experiments found candidates between \$0.5 and \$0.03 per-case cost. All Onboarding runs were performed in late July and early August, and model identifiers refer to OpenRouter's catalog at that time. 

\subsection{Measurement Reliability }
\label{sec:reliability}

\subsubsection{Judge Reliability}

The scoring in \Cref{eq:score} relies on an LLM judge (gpt-oss-120B at temperature 0.0) to determine whether the reasons for halting are correct or incorrect. To test its reliability, we sample 184 halted episodes that provided a reason: 8 per error class (\Cref{app:b_error_classes}), with 4 coming from baselines and 4 coming from varying Onboarded candidates. We labeled every sampled reason as incorrect or correct independently of the judge, and across all 184 episodes the two sets of verdicts were identical. Furthermore, every single halting reason was stated clearly and verbosely, indicating that the Onboarder is not simply finding a way to game the judge's verdict.

\subsubsection{Noise Estimation}
\label{sec:noise_estimation}

In addition, we test how noisy the experiments are and dissect variance into two sources. First, to measure evaluation noise, we perform a single Onboarding run for each cost goal (\Cref{sec:protocol}), and validate each resulting candidate three times. Then, to measure Onboarding noise, we repeat the entire Onboarding run at \(\beta = \$0.04\) three times, on an identical Onboarding dataset, and evaluate each resulting candidate once. The former measures the variance of the agentic evaluation task, whereas the latter measures the variance introduced by the Onboarding procedure in addition to that of the evaluation task. In \Cref{tab:noise}, we report the spread in \(S\), CH, MFA, and per-case cost for each noise estimation (\Cref{sec:task}). We also calculate the number of cases that flip outcome (success vs. failure) between at least two of the three repeats. The Onboarding run at the \(\$0.01\) cost goal was unable to produce a candidate under its \(1.5\beta\) cost ceiling, so it was not validated and is excluded from the results. In every case except for the second Onboarding repeat, no perfect form submissions occur, so their score equals exactly half of CH; the second Onboarding repeat saw 10/39 perfect form submissions. 

\begin{table}[tb]
\centering
\renewcommand{\arraystretch}{0.9}
\setlength{\tabcolsep}{8pt}
\begin{tabular}{llcccccc}
\toprule
Cost goal (\(\beta\))
    & Metric
    & Run 1
    & Run 2
    & Run 3
    & Std. Dev.
    & Range
    & Flips \\
\midrule

\multicolumn{8}{@{}l}{\ \textbf{Evaluation Noise} \vspace{0.4em}} \\

\multirow{4}{*}{\(\$0.10\)}
    & \(S\)       & 0.309 & 0.294 & 0.298 & 0.007 & 0.014 & \multirow{4}{*}{19} \\
    & MFA     & 0.423  & 0.487  & 0.455  & 0.032  & 0.064  & \\
    & CH    & 0.617  & 0.589  & 0.596  & 0.015  & 0.028 & \\
    & \$\!/case    & 0.086  & 0.083  & 0.083  & 0.002  & 0.004 & \\

\addlinespace
\addlinespace

\multirow{4}{*}{\(\$0.04\)}
    & \(S\)       & 0.323 & 0.340 & 0.326 & 0.009 & 0.018 & \multirow{4}{*}{24} \\
    & MFA    & 0.406  & 0.376  & 0.564  & 0.101  & 0.188  & \\
    & CH    & 0.645  & 0.681  & 0.652  & 0.019  & 0.035 & \\
    & \$\!/case    & 0.069  & 0.066  & 0.067  & 0.001  & 0.002 & \\

\addlinespace
\addlinespace

\(\$0.01\) & \multicolumn{6}{c}{\emph{No legal candidate produced.}} \\

\midrule

\multicolumn{8}{@{}l}{\ \textbf{Onboarding Noise} \vspace{0.4em}} \\

\multirow{4}{*}{\(\$0.04\)}
    & \(S\)       & 0.323 & 0.501 & 0.199 & 0.152 & 0.302 & \multirow{4}{*}{63} \\
    & MFA    & 0.406  & 0.917  & 0.041  & 0.440  & 0.876 & \\
    & CH    & 0.645  & 0.745  & 0.397  & 0.179  & 0.348 & \\
    & \$\!/case    & 0.069  & 0.032  & 0.052  & 0.019  & 0.037 & \\

\bottomrule
\end{tabular}
\caption{Spread of validation metrics (\Cref{sec:task}) across runs. Top: three validation repeats at varying cost goals. Bottom: three onboarding repeats at the \(\beta = \$0.04\) goal. Std. Dev. is the sample standard 
deviation calculated from the three repeats. Flips is the number of the 
180 cases whose outcome (failure vs. success) is not constant across all three repeats. The Onboarding run at cost goal \(\$0.01\) was unable to produce a cheap-enough candidate.}
\label{tab:noise}
\end{table}

In the evaluation noise estimation, the produced candidates only vaguely match their cost goals: the \(\beta = \$0.10\) run produces a candidate with a per-case cost of \(\$0.053\) during Onboarding and evaluates at a per-case cost of \(\$0.084 \pm \$0.004\), while the \(\beta = \$0.04\) run produces a candidate with a per-case cost of \(\$0.054\) during Onboarding and evaluates at a per-case cost of \(\$0.068 \pm \$0.002\). Notably, both candidates sit within the \(\beta = \$0.04\) run's \(1.5\beta = \$0.06\) ceiling during Onboarding, so we deem \(\beta = \$0.04\) the most informative goal and use it in all remaining experiments. Across both cost goals, the sample standard deviation of the score stays below 0.01, though about 20 cases (19 and 24) flip outcome.

\begin{figure}[tb]
\centering
\includegraphics[width=1\linewidth]{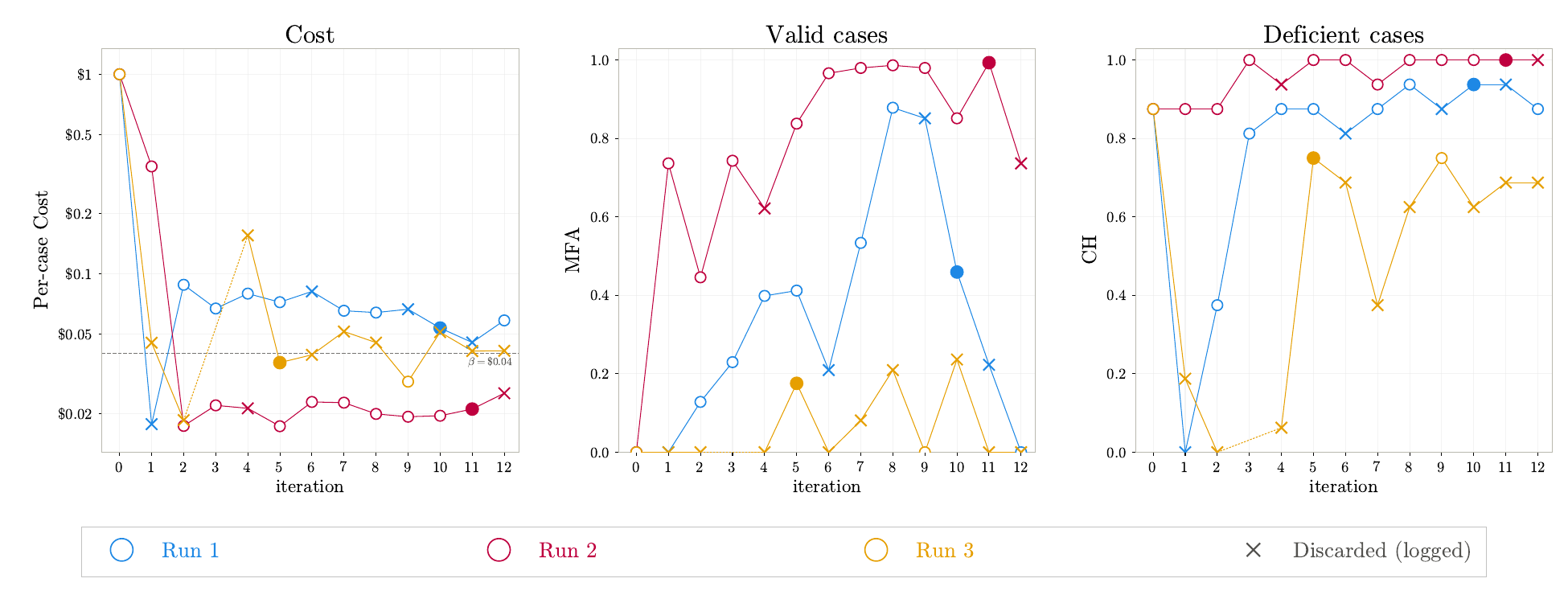}
\caption{Onboarding learning curves, displaying Per-Case Cost (left), MFA (middle), and CH (right) with respect to Onboarding iteration. Filled circles correspond to the nominated iteration, crosses correspond to iterations that were reverted and discarded, and open circles correspond to iterations that were kept, but not nominated. All reported calculations are performed on the Onboarding data, rather than the validation data. Iteration 3 of Run 3 was stopped by the Onboarder before scoring, so it is not plotted.}
\label{fig:paths}
\end{figure}

The Onboarding noise estimation is far greater than that of the evaluation noise: the sample standard deviation of \(S\) across Onboarding repeats (0.152) is roughly 20$\times$ its evaluation noise counterpart (0.007/0.009). \Cref{fig:paths} depicts this divergence more clearly, plotting the Onboarding per-case cost, MFA, and CH of these three runs as a function of iteration. Notably, the cost varies wildly -- in the validation results of \Cref{tab:noise}, the sample standard deviation of per-case cost (\$0.019) is about half of the cost goal, \(\$0.04\) -- and Onboarding-time MFA ranges from \(0\%\) to \(100\%\). Between the three Onboarding runs, there are 63 cases that flip outcomes.

\subsubsection{Flips}

\begin{figure}
    \centering
    \includegraphics[width=1\linewidth]{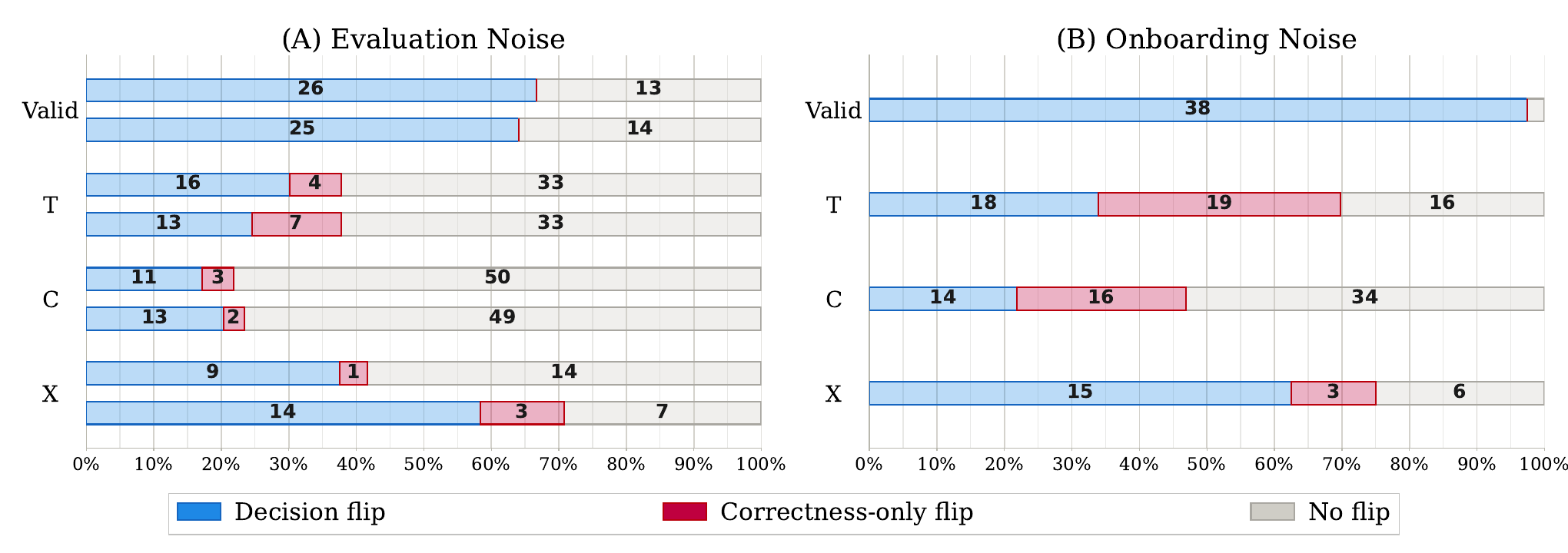}
    \caption{Decision flips of Onboarding runs by sample class type (\Cref{app:b_error_classes}). Left (A): Flips in evaluation noise repeats, top bars correspond to \(\beta = \$0.10\) and bottom bars correspond to \(\beta=\$0.04\). Right (B): Flips in Onboarding noise repeats.}
    \label{fig:flips}
\end{figure}

In all of these noise estimates, the calculated variance in the score understates how many cases flip outcome. Here, we perform an observational study over all \textit{behavior} flips: flips in \textit{any} of the Executor's behavior. We classify these behavior flips into two categories: ones where only the correctness of a response changes (i.e., going from incorrect halting to correct halting or imperfect submission to perfect submission) and ones where the decision also changes (i.e., going from halting on a valid case to imperfect/perfect submission or submitting a deficient case to incorrect/correct halting). In total, there are 70 and 77 behavior flips in the \(\beta = \$0.10\) and \(\beta = \$0.04\) evaluation repeats, respectively, and 123 in the Onboarding repeats. \Cref{fig:flips} plots the percentage of flips by sample class (\Cref{app:b_error_classes}). Seemingly, valid cases and cross-field contradictions induce the most prominent amount of noise. Moreover, on valid cases every flip is a submit/halt decision change -- the three Onboarding repeats submit 18, 37, and 2 of the 39 valid cases -- so the large MFA spread in \Cref{tab:noise} largely reflects whether a candidate submits forms at all, as opposed to how accurately it fills out forms.

\subsection{Comparison Across Lever Sets}
\label{sec:effectiveness}

To measure the impact of incorporating the model and harness into Onboarding, we compare runs that have access to the prompt+model+harness (PMH), only the prompt+model (PM), and only the prompt (P). Because it is infeasible to reduce costs from the baseline Gemini 3.5 Flash candidate solely through edits to the prompt, we initialize the P run from GPT-5.6 Luna, as it achieves a per-case cost of only \(\$0.027\) with the baseline configuration. We compare these three runs against a collection of baseline models, including Gemini 3.5 Flash and smaller models such as GPT-5.6 Luna, GPT-4.1 Nano, and Gemini 3.1 Flash-Lite. 

\begin{table}[tb]
\centering
\begin{singlespace}
\small
\setlength{\tabcolsep}{8pt}
\begin{tabular}{lcccccc}
\toprule
Candidate & \(S\) & MFA & CH & \shortstack{Perfect\\subs. (/39)} & \shortstack{Near-perfect\\subs. (/39)} & \$\!/case \\
\midrule
Gemini 3.5 Flash (stock) & 0.319 & 0.024 & 0.638 & 0 & 0 & 0.934 \\
GPT-5.6 Luna (stock) & 0.028 & 0.000 & 0.057 & 0 & 0 & 0.027 \\
Gemini 3.1 Flash-Lite (stock) & 0.000 & 0.393 & 0.000 & 0 & 0 & 0.041 \\
GPT-4.1 Nano (stock) & 0.000 & 0.000 & 0.000 & 0 & 0 & 0.042 \\
\midrule
PMH & 0.323 & 0.406 & 0.645 & 0 & 2 & 0.069 \\
PM & 0.325 & 0.773 & 0.624 & 1 & 7 & 0.036 \\
P (cheap init.) & 0.409 & 0.657 & 0.766 & 2 & 9 & 0.016 \\
\bottomrule
\end{tabular}
\end{singlespace}
\caption{Validation results (\Cref{sec:task}) of the Onboarded candidates at \(\beta = \$0.04\) and the stock baselines. Near-perfect submissions have exactly 36 of 37 fields correct.}
\label{tab:effectiveness}
\end{table}

All three Onboarded candidates match or exceed Gemini 3.5 Flash's score at a small fraction of its cost (\Cref{tab:effectiveness}). Furthermore, they improve its MFA from 0.024 to 0.406--0.773. Other cheap baselines perform far worse, with Gemini 3.1 Flash-Lite and GPT-4.1 Nano scoring 0.0 and GPT-5.6 Luna achieving a score of only 0.028.

However, due to the sizable variance in the Onboarding procedure and low sample size, we believe that comparisons between lever configurations (i.e., PMH vs. PM vs. P) cannot be interpreted in a meaningful context. While the P candidate (score 0.409) seemingly performs substantially stronger than the PM (0.325) and PMH (0.323) candidates, it performs worse than the second PMH repeat in \Cref{tab:noise} (0.501) and far better than the third one (0.199).

\subsection{ Generalization Analysis}
\label{sec:generalization}

\begin{figure}[tb]
\centering
  \includegraphics[width=\linewidth]{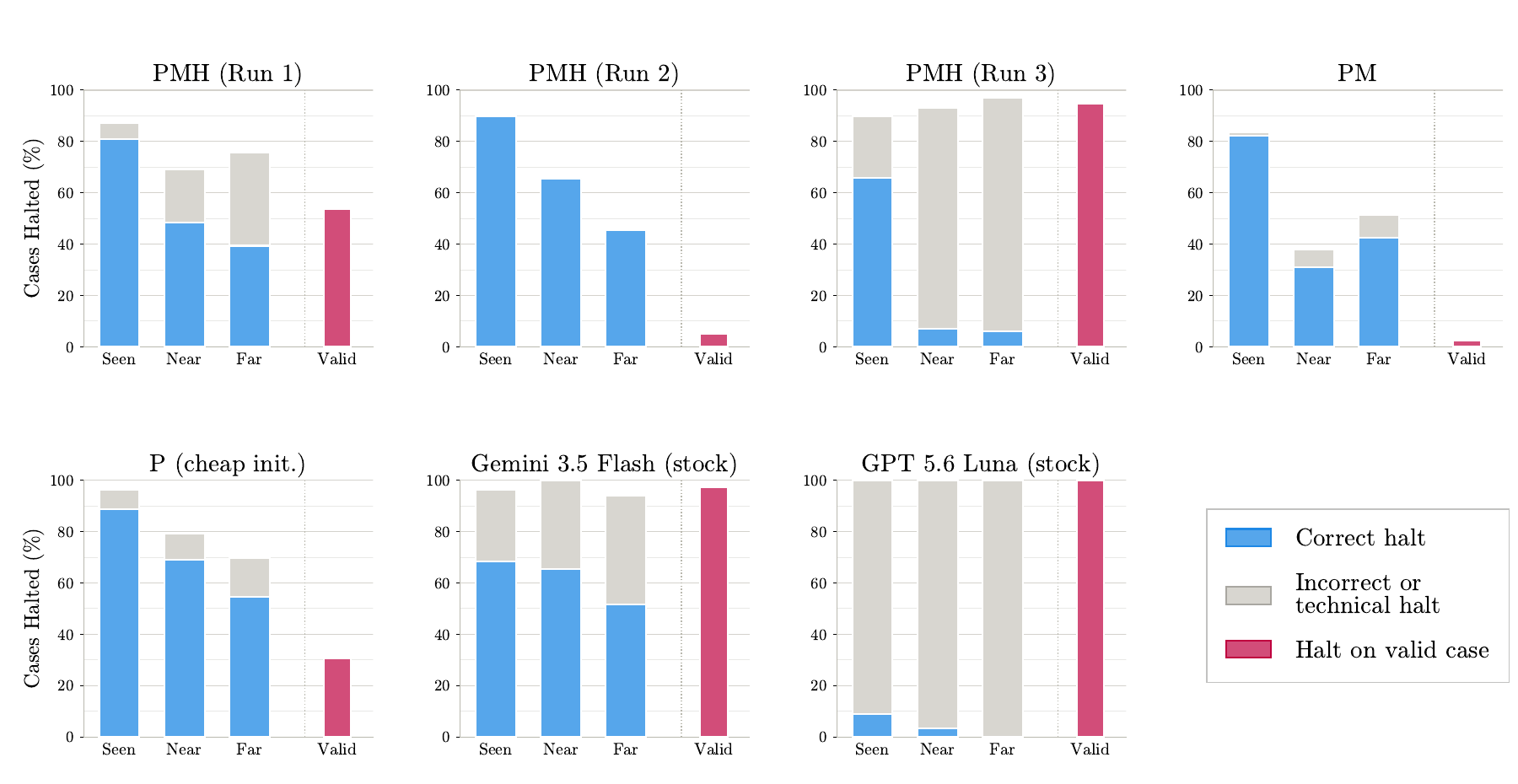}
  \caption{Halting behavior by coverage group (\Cref{sec:generalization}). For each stacked bar, the filled segment is the share halted for the correct reason, and the gray extension adds halts with an incorrect or technical reason, so the full bar is total halting. The separated bar in red reports halts on the 39 valid cases, which should have been submitted.}
  \label{fig:generalization}
\end{figure}

Finally, we perform an observational study of generalization to unseen error classes, splitting the error classes in the validation set (\Cref{app:b_error_classes}) into three coverage groups: (i) seen, which are represented in the Onboarding dataset; (ii) near transfer, which have a near sibling represented; and (iii) far transfer, which have no sibling represented. There are 29 near transfer cases, 33 far transfer cases, and 79 seen-class cases. We plot correct and incorrect halting among these three coverage groups, along with halting percentage on valid cases, among varying Onboarded candidates at the \(\beta = \$0.04\) cost goal and the baselines Gemini 3.5 Flash and GPT-5.6 Luna. \Cref{fig:generalization} shows the results for the three Onboarding repeats of \Cref{sec:reliability} and the PM and P Onboarding runs of \Cref{sec:effectiveness}.

While the Onboarding process can seemingly reduce the number of false-halts substantially (from 97.4\% of valid cases halted by the baseline Gemini 3.5 Flash to 53.8\% for the first PMH repeat), we are again unable to draw a meaningful conclusion about clean transfer from the data. By inspection, it seems as though much of the difference in success between error classes is inherent to case difficulty, as both Onboarded candidates and baselines swing widely across coverage groups.

\section{Discussion}

We close with the lessons we drew from building and testing AutoOnboarder. Beyond the validation results above, they draw on the Onboarder's per-iteration ledgers and session logs. 

\paragraph{Onboarding clearly shows promise. } Current experiments testify to the potential of AutoOnboarder: the best candidate exceeds the Gemini 3.5 Flash baseline in performance at a 29$\times$ lower cost and submits far more valid cases. Our design allows the Onboarder agent to effectively edit every component that shapes an Executor's behavior, while simultaneously reducing costs. The leakage-controlled two-machine setup is designed to prevent all data leakage, and post-hoc searches of Onboarder session logs and API logs found no attempts to bypass data protection.

\paragraph{However, Onboarding exhibits excessively large variance.} Experiments are incredibly noisy -- repeat runs' scores range from 0.199 to 0.501, per-case cost from \$0.032 to \$0.069, and valid submissions from 2 to 37 (of 39). Much of the Onboarding spread seems to trace back to Onboarder's environment and search space; the next three lessons take up each of these in turn.

\paragraph{The model lever needs a curated search space.} At the cost scales we attempt to cut to, OpenRouter models are often unreliable. Many affordable models on OpenRouter, such as Mistral Small, Llama 4 Scout, and Gemini 2.5 Flash, are unable to conform to browser-use's output schema; others, like Qwen3 VL 235B, fall back to providers that lack a required feature like image input or are heavily rate limited (HTTP 429). Across six runs with access to the model lever, 24 iterations made changes to the foundation model, and 15 of them ended as crashes or discards. Each dead end consumes a full pass over the Onboarding set and one of only 12 iterations, which contributed significantly to divergence between Onboarding runs. Of the three Onboarding repeats at \(\beta = \$0.04\), failed model swaps accounted for the early discarded iterations in the two non-optimal runs. We therefore believe a curated, wider model catalog is a necessity for Onboarding. Furthermore, crafting search spaces that incorporate distilled, quantized, or otherwise compressed versions of the same foundation LLMs would likely improve onboarding quality by granting more fine-grained control in the model lever. 

\paragraph{The Onboarder needs a structured interface for thought.} In our testing, we notice that AutoOnboarder tends to be quite sensitive and unpredictable in regard to the wording of its contract. Many behavioral instructions are completely ignored or taken to extremes. It often picks models by name recognition and ignores the benchmark data from its helper script, declares its task complete or impossible with iterations left unspent, and only loosely meets cost goals (for example, the \(\beta = \$0.10\) run ended Onboarding with a per-case cost of \(\$0.053\)). Furthermore, this obstinate behavior sometimes leads the Onboarder to give up on form submission entirely, and we believe it may be the root of a significant portion of the Onboarding variance. 

Another failure mode we noticed in our experimentation is that the harness lever is used sparingly. In all of our experiments, the Onboarder makes tweaks to the harness lever in at most two iterations, even after we explicitly instruct it to use all of its levers. Examining reasoning traces of the Onboarder does not provide a clear picture of why this is the case; we read this as a limitation of our implementation rather than of the value of edits to the harness. Unlike meta-agents such as ADAS~\cite{shengran_adas_2024} and the Darwin Gödel Machine~\cite{zhang_darwin_2025}, which are given hundreds of iterations to rewrite their harnesses, AutoOnboarder is only given 12. With so few iterations and so strict a cost budget, it is possible that the Onboarder  simply did not have enough room to experiment with the harness lever sufficiently.

Together, these observations suggest that the Onboarder's free-form, contract-driven interface is itself a bottleneck. Reconciling these issues likely requires a more robust interface for the Onboarder that provides it more tools and forces it to consolidate its reasoning, much like that designed for Executor agents in SWE-Agent~\cite{yang_swe_2024} and AgentOccam~\cite{yang_occam_2024}. Such an interface would enforce the Onboarder's rules rather than leave them open to its interpretation of a written contract. 

\paragraph{The Onboarder requires controlled, not absent, internet access.} To keep the system secure, the Onboarder has no network access. This isolation, though, also limits what the Onboarder can teach. Perfect submissions require exact ICD-10 codes (a clinical diagnosis standard) drawn from a relatively wide pool. Without access to the internet, the Onboarder often explicitly acknowledges that it has no way to package an ICD database for the Executor. We therefore believe that much of the residual imperfection may be a missing-tooling problem, rather than one baked into the prompt, model, or harness. However, unrestricted internet access would make it easy for an Onboarder to reward hack, for example by seeking out answers to specific cases instead of teaching general skills. We therefore argue for controlled access: an allowlisted, logged, read-only channel, such as a proxy or curated mirrors of resources like ICD-10 tables and model and provider documentation, that sits inside the existing boundaries of HMAC-signed submissions, filesystem permissions, and validation data isolated on VM2.

 The primary operational limitations of this study stem from the time and cost of the experiments, which reduce the number of repeat experiments we can perform: a single case can take over 30 minutes, and evaluating a model like Gemini 3.5 Flash on the 200-case set takes around \(\$190\). 

Interestingly, in our experiments, Gemini 3.5 Flash submits only 6 of 180 cases and wrongly withholds 38 of the 39 valid ones, which conflicts with the automation bias claim of Nie et al.~\cite{nie_hard_2026}. We hypothesize that the differences lie primarily in closed-model drift, whereby commercial LLM providers silently change LLM behavior over time. This is also a salient focus in our study: as with any research on closed LLMs, it is possible that some findings in this work do not generalize to newer or tweaked models. 

\section{Conclusion}

In this work, we set out to develop an autonomous system that would adapt an agent and its harness to a task, thereby recovering the losses in critical thinking skills documented by Nie et al.~\cite{nie_hard_2026}. Current experiments testify to the potential of AutoOnboarder in reducing Executor agents' costs while maintaining or improving their scores. However, current experiments are also prone to large variance and subpar performance.

In the future, we plan a more robust system where the Onboarder has an interface that forces more structured thinking and includes all the tools it needs, while also balancing this freedom with data security. We hypothesize that such a system would both reduce the large variance of Onboarding runs and contribute to better Executor performance.

\section{Key Takeaways}

In this work, we show that Onboarding has the potential to adapt AI agents to be more reliable and cost-effective, making them more practical in production. In the future, agentic systems like AutoOnboarder can drive the creation of small, cheap agents that take high-volume busy work, like prior authorization forms, off of human hands. However, this work also demonstrates the difficulties in creating such a system: the same Onboarding procedure, run twice, can produce very different agents, and no candidate is yet dependable enough for real clinical use. Moving forward, we need to develop a more repeatable, robust method for holistic agent-optimizing systems that prioritizes consolidated thinking and more closely balances security with access to information.

\section*{Acknowledgements}
\addcontentsline{toc}{section}{Acknowledgements}

We thank Mengshu Nie for building the prior-authorization sandbox that this work extends, and the rest of CliGenAILab, especially Dr.\ Kamalakkannan Ravi for his help in proposing this project. Finally, we thank the Center for Excellence in Education, the Research Science Institute, and the Massachusetts Institute of Technology for making this research possible.

\bibliographystyle{style}
\bibliography{biblio}

\appendix
\gdef\thesection{Appendix \Alph{section}}

\appendix

\section{Error Classes}
\label[appendix]{app:b_error_classes}

Each case corresponds to a single name, which is passed to the Executor inside of its prompt. The dataset in this paper expands the dataset of \cite{nie_hard_2026} from five deficiency signatures to 23 (\Cref{sec:task}). Percentages below are shares of all 180 validation cases. Classes P and Dsec require submission; the 23 deficient classes require withholding and reporting the specific problem. Error classes can be split into data-integrity (T) issues in transcription fields, clinical necessity (C) issues in the clinical note, and cross-field issues (X) that have contradictions between transcribed fields and the clinical note. Clinical notes were produced with OpenAI's GPT-5.6 Terra with no reasoning. We define siblings as classes in the same subgroup. For example, T3a and T3b are siblings, while T6 has no siblings.

\begin{table}[H]
\centering
\begin{singlespace}
\scriptsize
\setlength{\tabcolsep}{4pt}
\renewcommand{\arraystretch}{1.2}
\begin{tabular}{@{}p{0.2\linewidth}p{0.26\linewidth}p{0.52\linewidth}@{}}
\toprule
\textbf{Group} & \textbf{Class: cases (\% of dataset)} & \textbf{Definition} \\
\midrule
\textbf{Valid} & \textbf{P}: 27 (15.0\%) & Valid, well-supported request. \\[0.85ex]
39 (21.7\%) & \textbf{Dsec}: 12 (6.7\%) & Valid request with unrelated secondary details; should still submit. \\
\midrule
\textbf{Data-integrity} (T) & \textbf{T1a}: 4 (2.2\%); \textbf{T1b}: 5 (2.8\%) & Subscriber listed as a parent is implausibly close in age to the patient, or born after the patient. \\[0.85ex]
53 (29.4\%) & \textbf{T2}: 6 (3.3\%) & Specimen collection date precedes the patient's date of birth. \\[0.85ex]
& \textbf{T3a}: 6 (3.3\%); \textbf{T3b}: 6 (3.3\%) & Two patients share an identical name (and, for T3b, date of birth), making the case ambiguous. \\[0.85ex]
& \textbf{T4a}: 5 (2.8\%); \textbf{T4b}: 5 (2.8\%) & Provider or lab NPI, or provider phone or fax, is not valid. \\[0.85ex]
& \textbf{T5a}: 2 (1.1\%); \textbf{T5b}: 2 (1.1\%); \textbf{T5c}: 2 (1.1\%); \textbf{T5d}: 2 (1.1\%); \textbf{T5e}: 2 (1.1\%); \textbf{T5f}: 2 (1.1\%) & Required element is missing: provider NPI, lab NPI, provider address, provider phone, subscriber date of birth, or collection date, respectively. \\[0.85ex]
& \textbf{T6}: 4 (2.2\%) & Member ID is malformed. \\
\midrule
\textbf{Clinical necessity} (C) & \textbf{C1}: 13 (7.2\%) & Acquired or environmental issue with no heritable indication. \\[0.85ex]
& \textbf{C2}: 11 (6.1\%) & Isolated, common, non-syndromic finding. \\[0.85ex]
64 (35.6\%) & \textbf{C3}: 11 (6.1\%) & Screening or carrier indication in an asymptomatic patient. \\[0.85ex]
& \textbf{C4}: 11 (6.1\%) & Causative diagnosis already made, so sequencing is redundant. \\[0.85ex]
& \textbf{C5a}: 9 (5.0\%); \textbf{C5b}: 9 (5.0\%) & Sequencing already performed, or prior targeted testing already diagnostic. \\
\midrule
\textbf{Cross-field} (X) & \textbf{X1}: 8 (4.4\%) & Recorded sex contradicts the sex described in the note. \\[0.85ex]
& \textbf{X3}: 8 (4.4\%) & Collection date precedes the prior test described in the note. \\[0.85ex]
24 (13.3\%) & \textbf{X4}: 8 (4.4\%) & Broad sequencing is ordered where a targeted single-gene test is the indicated next step. \\
\midrule
\textbf{Total} & \textbf{180 (100\%)} & \\
\bottomrule
\end{tabular}
\end{singlespace}
\caption{Validation-set sample classes and prevalence. The set contains 39 valid cases (21.7\%) and 141 deficient cases (78.3\%); every subclass has at least two cases.}
\label{tab:error_classes}
\end{table}

\end{document}